\documentclass{article} 
\usepackage{iclr2023_conference_tinypaper,times}

\usepackage{hyperref}
\usepackage{url}
\usepackage{graphicx}
\usepackage{amsmath}
\usepackage{amssymb}
\usepackage{booktabs}
\usepackage{multirow}
\usepackage{algorithm}
\usepackage{dsfont}
\usepackage{tikz} 
\usetikzlibrary{arrows}
\usepackage{bbm}
\usepackage{wrapfig}
\usepackage{svg}
\usepackage{multirow}
\usepackage{subfig}
\usepackage[font=small,labelfont=bf]{caption}

\title{Averaging Rate Scheduler for Decentralized Learning on Heterogeneous Data}

\author{Sai Aparna Aketi, Sakshi Choudhary \& Kaushik Roy \\
Electrical and Computer Engineering\\
Purdue University\\
West Lafayette, IN 47906, USA \\
\texttt{\{saketi,choudh23,kaushik\}@purdue.edu} \\
}

\iclrfinalcopy 
\begin{document}

\maketitle

\begin{abstract}
State-of-the-art decentralized learning algorithms typically require the data distribution to be Independent and Identically Distributed (IID). However, in practical scenarios, the data distribution across the agents can have significant heterogeneity. In this work, we propose averaging rate scheduling as a simple yet effective way to reduce the impact of heterogeneity in decentralized learning. Our experiments illustrate the superiority of the proposed method ($\sim 3\%$ improvement in test accuracy) compared to the conventional approach of employing a constant averaging rate.
\end{abstract}

\section{Introduction}

Decentralized Learning, a subset of Federated Learning, offers many advantages over the traditional centralized approach in core aspects such as data privacy, fault tolerance, and scalability without the need for a central server.
Research has shown that decentralized learning algorithms can perform comparable to centralized algorithms on benchmark datasets \citep{d-psgd} under the assumption of IID data. 
Recently, there have been several efforts \citep{ngm, relaysgd} that make algorithmic changes to bridge the performance gap between IID and non-IID data for decentralized setups. 
In this work, we explore an orthogonal direction of scheduling the averaging rate to handle heterogeneous data in decentralized setups.

The averaging rate or consensus step size \citep{deepsqueeze, choco-sgd} is a hyper-parameter introduced in decentralized algorithms with communication compression to control the rate at which the model parameters across the neighbors are averaged. The value of the averaging rate depends on the compression ratio and remains constant throughout the training. The higher the compression ratio, the lower the value of the averaging rate. 
However, the averaging rate is usually ignored when there is no communication compression and is set to a constant value of one. We hypothesize that the averaging rate is crucial in decentralized setups with heterogeneous data. In this paper, we validate two significant claims: 1. Tuning the averaging rate is critical for decentralized learning on heterogeneous data (Figure.~\ref{fig:ar_tune}), and 2. Scheduling the averaging rate during training helps improve the performance of decentralized learning on heterogeneous data (Table.~\ref{table:models}).

\section{Background and Related Work}
The main goal of decentralized learning is to learn a global model utilizing data stored locally across $n$ agents connected via a sparse graph.
Traditional decentralized algorithms \citep{d-psgd} include three steps at every iteration - (a) Local SGD update, (b) Communication, and (c) Gossip averaging.
Algorithm.~\ref{alg:dl} in the appendix describes the baseline Decentralized Stochastic Gradient Descent (DSGD).
Each agent computes the gradients using local data and performs an SGD update. Then, the updated parameters are communicated to the neighbors. Finally, in the gossip averaging step, the received model parameters of the neighborhood are averaged using the mixing weights. 

Several works in the literature aim to improve the performance of decentralized learning for heterogeneous data. Some of these works include tracking-based methods \citep{gt, mt, gut}, cross-gradients based methods \citep{ngm, cga}, momentum-based methods \citep{qgm}, changing the mixing matrix \citep{dandi2022data} etc. However, all of these works use a constant averaging rate throughout the training and don't explore the impact of scheduling this hyper-parameter.

\section{Averaging Rate Scheduler}

In decentralized learning, \emph{averaging rate} is a hyper-parameter that indicates the rate at which the model parameters are averaged across the neighborhood. Typically, the averaging rate is constant, often set to one. However, in a heterogeneous data setting, the model parameters across the neighborhood exhibit significant differences during the initial training phase.
In this scenario, a high averaging rate (set to one) can disrupt local training by adding large gossip error (refer \ref{apx:dl} for definition) to local model parameters.
To circumvent this, we propose to employ an Averaging Rate Scheduler (ARS) for decentralized learning. 
The proposed scheduler initializes the averaging rate to a tuned lower value and gradually increases it to one over training. 
Note that, unlike the learning rate, the averaging rate increases with time. 
Similar to the learning rate schedulers, one can explore schedulers such as exponential, step, multi-step, cosine, etc. 
The scheduler scales the averaging rate by a factor known as \emph{growth rate} after every $k$ epochs.  
The value of the growth rate and $k$ depend on the type of scheduler. 
Through our exhaustive experiments, we demonstrate that the ARS improves the performance of decentralized learning on heterogeneous data. 

\section{Result and Conclusion}

\begin{table}
\begin{minipage}{0.46\linewidth}
		\centering
		\includegraphics[width=50mm]{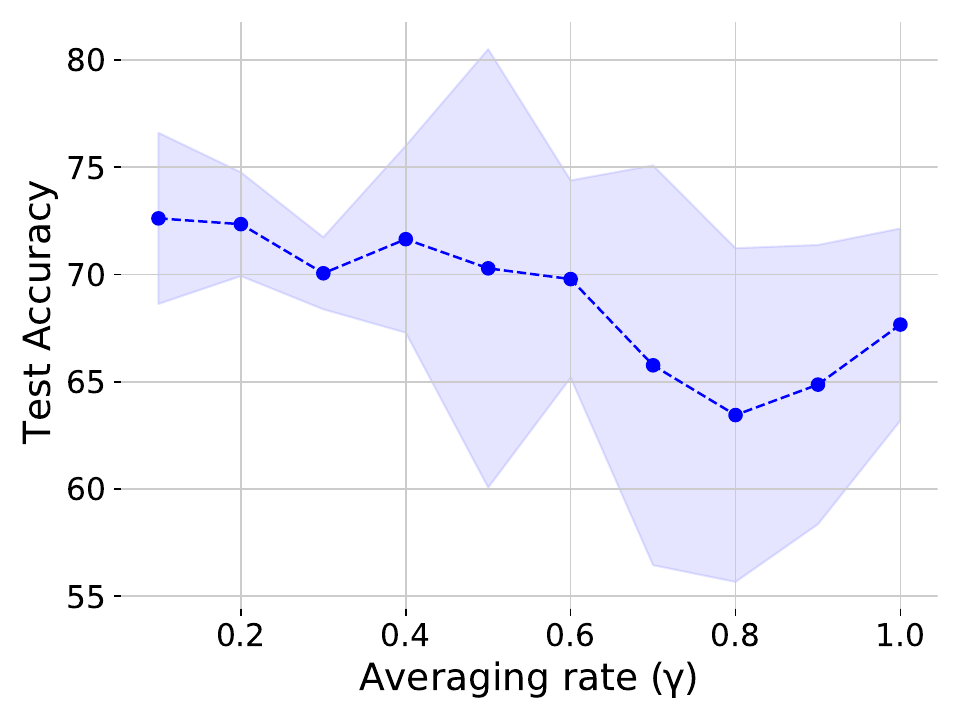}
		\captionof{figure}{Variation of test accuracy with averaging rate (constant during training) during training of CIFAR-10 on ResNet-20 over a ring graph of 16 agents with $\alpha=0.01$.}
		\label{fig:ar_tune}
	\end{minipage} \hfill
	\begin{minipage}{0.52\linewidth}
 \caption{Different Schedulers used in training of CIFAR-10 on ResNet-20 over a ring topology of 16 agents.}
 \vspace{-2mm}
		\label{table:schedulers}
		\centering
		\begin{tabular}{lc}
			\toprule
			Scheduler &  Test Accuracy \\ 
			\midrule
            Step & $ 70.46 \pm 5.08 $  \\
            Exponential & $\mathbf{71.19} \pm 3.66$ \\
            Cosine & $70.81 \pm 5.59$  \\
			\bottomrule
		\end{tabular}
        \vspace{2mm}
		\caption{Impact of exponential Averaging Rate Scheduler (ARS) on CIFAR-10 dataset trained over a ring topology with 16 agents using DSGDm-N algorithm.}
        \vspace{-2mm}
		\label{table:models}
		\centering
		\begin{tabular}{lcc}
			\toprule
			Model & w/o ARS & w/ ARS\\ 
			\midrule
            ResNet-20 & $67.67 \pm 4.48$ & $\mathbf{71.19} \pm 3.66$ \\
            VGG-11 & $68.38 \pm 0.83$ & $\mathbf{71.54} \pm 1.60$ \\ 
			\bottomrule
		\end{tabular}
	\end{minipage}
\end{table}

We evaluate the effectiveness of the proposed ARS on a diverse set of datasets and models over a ring topology \footnote{The PyTorch code is available at \url{https://github.com/aparna-aketi/ARS}}.
Following \citep{qgm}, we use a Dirichlet distribution-based partitioning with a parameter $\alpha$ quantifying the heterogeneity between the data distributions across agents. We set $\alpha=0.01$ as it corresponds to a setting with high heterogeneity. 
The hyper-parameters at all the agents are synchronized at the beginning of the training (Appendix.~\ref{apx:hp}) and all the agents have the same ARS.
The experiments are run for three seeds and the averaged test accuracy of the consensus model with standard deviation is presented.
Figure.~\ref{fig:ar_tune} shows that tuning the averaging rate instead of using the default value of 1 improves the performance. Note that in Figure.~\ref{fig:ar_tune} we use a constant averaging rate throughout the training. Table.~\ref{table:schedulers} shows the impact of various types of ARS on the CIFAR-10 dataset.
We observe that the exponential scheduler gives the best accuracy. Table.~\ref{table:models} illustrates the gain in performance obtained by the addition of an exponential ARS. We observe an accuracy improvement of $\mathbf{3.5}\%$ for ResNet-20 and $\mathbf{3.2}\%$ for VGG-11 with a heterogeneous distribution of the CIFAR-10 dataset across 16 agents. The implementation details of the scheduler and the results on other datasets and models can be found in the Appendix.~\ref{apx:hp}, \ref{apx:results} respectively.

In summary, this paper presents the hypothesis that the averaging rate plays a crucial role in decentralized learning on heterogeneous data, and employing an averaging rate scheduler can improve the performance of such setups. Scheduling the averaging rate (for heterogeneous data) helps to reduce the impact of neighbors' updates in the initial exploratory phase of training when the variation in model updates across the agents is high.  
Our experimental results corroborate the proposed hypothesis. The potential future direction for this work is to study the impact of ARS on various decentralized learning algorithms (refer Appendix.~\ref{apx:future}) designed for heterogeneous data such as QG-DGSDm \citep{qgm}, NGM \citep{ngm}, etc. 

\section*{URM Statement}
The authors acknowledge that at least one key author of this work meets the URM criteria of the ICLR 2024 Tiny Papers Track.

\section*{Acknowledgements}
This work was supported by the Center for the Co-Design of Cognitive Systems (COCOSYS), a DARPA-sponsored JUMP center of Semiconductor Research Corporation (SRC).

\bibliography{iclr2023_conference_tinypaper}
\bibliographystyle{iclr2023_conference_tinypaper}

\newpage

\appendix
\section{Appendix}
\subsection{Decentralized Learning}
\label{apx:dl}
The main goal of decentralized learning is to learn a global model using the information extracted from the locally stored data across $n$ edge agents while maintaining privacy constraints. In particular, we solve the optimization problem of minimizing global loss function $f(x)$ distributed across $n$ agents as given in equation.~\ref{eq:1}. 
Here, $f_i$ is a local loss function (for example, cross-entropy loss) computed in terms of the data sampled ($d_i$) from the local dataset $D_i$ at agent $i$ with model parameters $x_i$.
\begin{equation}
\label{eq:1}
\begin{split}
    \min \limits_{x \in \mathbb{R}^d} f(x) &= \frac{1}{n}\sum_{i=1}^n f_i(x), \\
    and \hspace{2mm} f_i(x) &= \mathbb{E}_{d_i \in D_i}[F_i(x; d_i)] \hspace{2mm} \forall i
\end{split}
\end{equation}
This is typically achieved by combining stochastic gradient descent \citep{sgd} with global consensus-based gossip averaging \citep{gossip}. 
The communication topology in this setup is modeled as a graph $G = ([n], E)$. An edge $\{i,j\}$ is present in $E$ if and only if is a communication link between agents $i$ and $j$ to exchange the messages directly. 
$\mathcal{N}_i$ is used to represent the neighbors of $i$ including itself. 
It is assumed that the graph $G$ is strongly connected with self-loops.
The adjacency matrix of the graph $G$ is referred to as a mixing matrix $W$ where $w_{ij}$ is the weight associated with the edge $\{i,j\}$. Note that, weight $0$ indicates the absence of a direct edge between the agents. 
We assume that the mixing matrix is doubly stochastic and symmetric, similar to all previous works in decentralized learning \citep{d-psgd, qgm}. 

\begin{algorithm}[ht]
\textbf{Input:} Each agent $i \in [1,n]$ initializes model weights $x_i^{(0)}$, step size $\eta$, averaging rate $\gamma$, and  mixing matrix $W=[w_{ij}]_{i,j \in [1,n]}$, $\mathcal{N}_i$ represents neighbors of agent $i$ including itself.\\

Each agent simultaneously implements the 
T\text{\scriptsize RAIN}( ) procedure\\
1.  \textbf{procedure} T\text{\scriptsize RAIN}( ) \\
2.  \hspace{4mm}\textbf{for} t=$0,1,\hdots,T-1$ \textbf{do}\\
3.  \hspace*{8mm}$d_i^{t} \sim D_i$ \hfill \textcolor{gray!80}{// mini-batch sampling}\\
4.  \hspace*{8mm}$g_{i}^{t}=\nabla_x F_i(d_i^{t}; x_i^{t}) $ \hfill \textcolor{gray!80}{// Compute local gradients}\\
5.  \hspace*{8mm}$x_{i}^{t+\frac{1}{2}}=x_i^{t}-\eta g_i^{t}$ \hfill \textcolor{gray!80}{// Local SGD step}\\
6.  \hspace*{8mm}S\text{\scriptsize END}($x_{i}^{t+\frac{1}{2}}$) 
 \hspace{2mm} R\text{\scriptsize ECEIVE}($x_{j}^{t+\frac{1}{2}}$) \hfill \textcolor{gray!80}{// Communicate parameters across neighbors}\\
7.  \hspace*{8mm}$x_i^{(t+1)}=x_{i}^{t+\frac{1}{2}} + \gamma \sum_{j\in \mathcal{N}_i} w_{ij} (x_{j}^{t+\frac{1}{2}}-x_{i}^{t+\frac{1}{2}})$ \hfill \textcolor{gray!80}{// Gossip averaging step}\\
8.  \textbf{return}
\caption{Decentralized Learning with \textit{DSGD} \citep{d-psgd}}
\label{alg:dl}
\end{algorithm}

Algorithm.~\ref{alg:dl} describes the flow of Decentralized Stochastic Gradient Descent (DSGD) which is the baseline algorithm. DSGD has three main steps at every time step - (a) Local update, (b) Communication, and (c) Gossip averaging. At every iteration, each agent computes the gradients using local data and updates its model parameters as shown in line 6 of Alg.~\ref{alg:dl}. Then these updated model parameters are communicated to the neighbors as shown in line 7 of Alg.~\ref{alg:dl}. Finally, in the gossip averaging step, the local model parameters are averaged with the received model parameters of the neighbors using the mixing weights (shown in line 8 of Alg.~\ref{alg:dl}).

\textbf{Gossip error:} The term $\sum_{j\in \mathcal{N}_i} w_{ij} (x_{j}^{t+\frac{1}{2}}-x_{i}^{t+\frac{1}{2}})$ in line 7 of Alg.~\ref{alg:dl} is referred to as the \emph{gossip error} for agent $i$. The gossip error is scaled by the averaging rate before adding to the local model parameters. 

\subsection{Experimental Details}
\label{apx:hp}
For the decentralized setup, we use an undirected ring topology with a uniform mixing matrix. The undirected ring topology for any graph size has 3 peers per agent including itself and each edge has a mixing weight of $\frac{1}{3}$. We use the Dirichlet distribution \citep{qgm} to generate heterogeneous data distribution (skewed label partition) across the agents. The created data partition across the agents is fixed, non-overlapping, and never shuffled across agents during the training. All the experiments have the degree of heterogeneity set to a value of $\alpha=0.01$.  
DSGDm-N indicates the Decentralized Parallel Stochastic Gradient Descent algorithm \citep{d-psgd} with Nesterov momentum. All the experiments presented in the paper utilize D-PSGD as the decentralized learning algorithm.

\subsubsection{Datasets}

We use the following datasets for the image classification task. 

\textbf{CIFAR-10:} 
CIFAR-10 \citep{cifar} is an image classification dataset with 10 classes. The image samples are colored (3 input channels) and have a resolution of $32 \times 32$. 
There are $50,000$ training samples with $5000$ samples per class and $10,000$ test samples with $1000$ samples per class.

\textbf{CIFAR-100:} 
CIFAR-100 \citep{cifar} is an image classification dataset with 100 classes. The image samples are colored (3 input channels) and have a resolution of $32 \times 32$. There are $50,000$ training samples with $500$ samples per class and $10,000$ test samples with $100$ samples per class. 

\textbf{Fashion MNIST:}
Fashion MNIST \citep{fmnist} is an image classification dataset with 10 classes. The image samples are in greyscale (1 input channel) and have a resolution of $28 \times 28$. There are $60,000$ training samples with $6000$ samples per class and $10,000$ test samples with $1000$ samples per class.

\textbf{Imagenette:}
Imagenette \citep{imagenette} is a 10-class subset of the ImageNet dataset. The image samples are colored (3 input channels) and have a resolution of $224 \times 224$. There are $9469$ training samples with roughly $950$ samples per class and $3925$ test samples. 

\subsubsection{Model architectures}

We replace ReLU+BatchNorm layers of all the model architectures with EvoNorm-S0 as it was shown to be better suited for decentralized learning over non-IID data \citep{qgm}.
The details of the model architectures are given below. 

\textbf{ResNet-20:} For ResNet-20 \citep{resnet}, we use the standard architecture with $0.27M$ trainable parameters except that BatchNorm+ReLU layers are replaced by EvoNorm-S0.

\textbf{VGG-11:} We modify the standard VGG-11 \citep{vgg} architecture by reducing the number of filters in each convolutional layer by $4\times$ and using only one dense layer with 128 units. 
Each convolutional layer is followed by EvoNorm-S0 as the activation-normalization layer. VGG-11 has $0.58M$ trainable parameters.

\textbf{LeNet-5:} For LeNet-5 \citep{lenet}, we use the standard architecture with $61,706$ trainable parameters.

\textbf{MobileNet-V2:} We use the the standard MobileNet-V2 \citep{mobilnetv2} architecture used for CIFAR dataset with $2.3M$ parameters except that BatchNorm+ReLU layers are replaced by EvoNorm-S0.

\subsubsection{Hyper-Parameters}

All the experiments for CIFAR-10 are run for 300 epochs whereas the experiments for other datasets are run for 100 epochs. The initial learning rate is set to 0.1 for CIFAR datasets and 0.01 for Fashion MNIST and ImageNette. The learning rate is decayed by 10 at epochs 150 and 180 for CIFAR-10 and at 50 and 75 for the remaining datasets. We use Nesterov momentum with a momentum coefficient of 0.9, weight decay of $0.0001$, and a mini-batch size of 32 per agent. 
The experiments that do not employ ARS have the averaging rate set to a constant value of 1.
The results are averaged over three seeds and the average test accuracy and standard deviation of the consensus model are presented. The consensus model is obtained by averaging the model parameters of all the agents at the end of the training using the all-reduce method. 

\subsubsection{Averaging Rate Scheduler Details} 

For Table.~\ref{table:schedulers}, we use an initial averaging rate of 0.08, and the averaging rate is gradually increased to 1 during training. For the exponential scheduler, we use a growth rate of 1.01, and $k$ is set to 1 which means that the averaging rate is multiplied by 1.01 after every epoch. 
For the step scheduler, we use a growth rate of 1.09, and $k$ is set to 10 which means that the averaging rate is multiplied by 1.09 after every 10 epochs. 
For both of these schedulers, we clip the averaging rate at 1 ensuring that it never takes a value greater than 1. The cosine scheduler utilizes Equation.~\ref{eq:cosine}. Here, we use $\gamma_0$ as 0.08 and $T_{max}$ is 300, and $k$ is set to 1 implying that the averaging rate is updated according to equation.~\ref{eq:cosine} after every epoch. 

\begin{equation}
\label{eq:cosine}
\begin{split}
    \gamma_t = \gamma_{0} + \frac{1}{2}(1-\gamma_{0})(1-cos\big(\frac{t}{T_{max}}\pi \big))
\end{split}
\end{equation}

For Table.~\ref{table:models}, \ref{table:agents}, we use an exponential averaging rate for both ResNet-10 and VGG-11 models. In the case of ResNet-20 experiments, the initial averaging rate is set to 0.08 and the growth rate is set to 1.01. In the case of VGG-11, the initial averaging rate is set to 0.09 and the growth rate is set to 1.01. The hyper-parameter pair values (initial averaging rate, growth rate) for Table.~\ref{table:datasets} are Fashion-MNIST: (0.1, 1.03), CIFAR-100: (0.5, 1.02), and ImageNette: (0.1, 1.03).

\subsection{Additional Results}
\label{apx:results}

We provide additional results on various graph sizes and datasets to show the 
generalizability and scalability of the ARS. Table.~\ref{table:agents} shows the results on varying graph (ring) sizes from 16 to 48 agents for training CIFAR-10 dataset on ResNet-20 architecture. We observe that D-PSGD with ARS achieves better performance with $3.5-6.9 \%$ improvement in test accuracy compared to D-PSGD without ARS. 
\begin{figure}[h]
  \centering
  \subfloat[CIFAR-10 trained on a 5-layered CNN]{\includegraphics[width=0.4\textwidth]{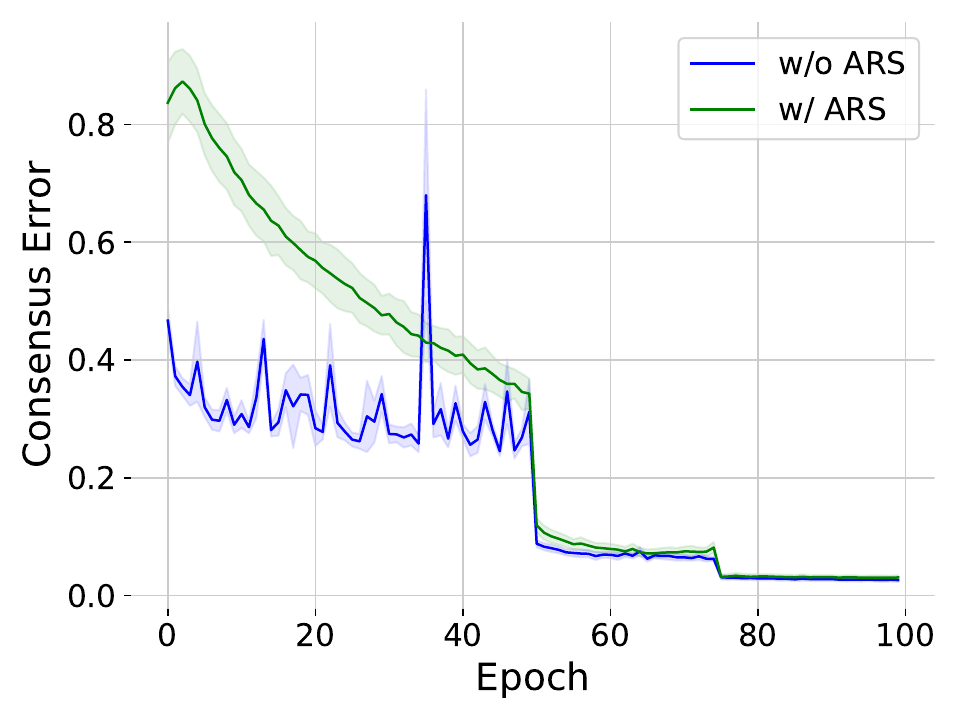}\label{fig:ce_cf10}}
  \hspace{4mm}
  \subfloat[Fashion-MNIST trained on LeNet-5]{\includegraphics[width=0.4\textwidth]{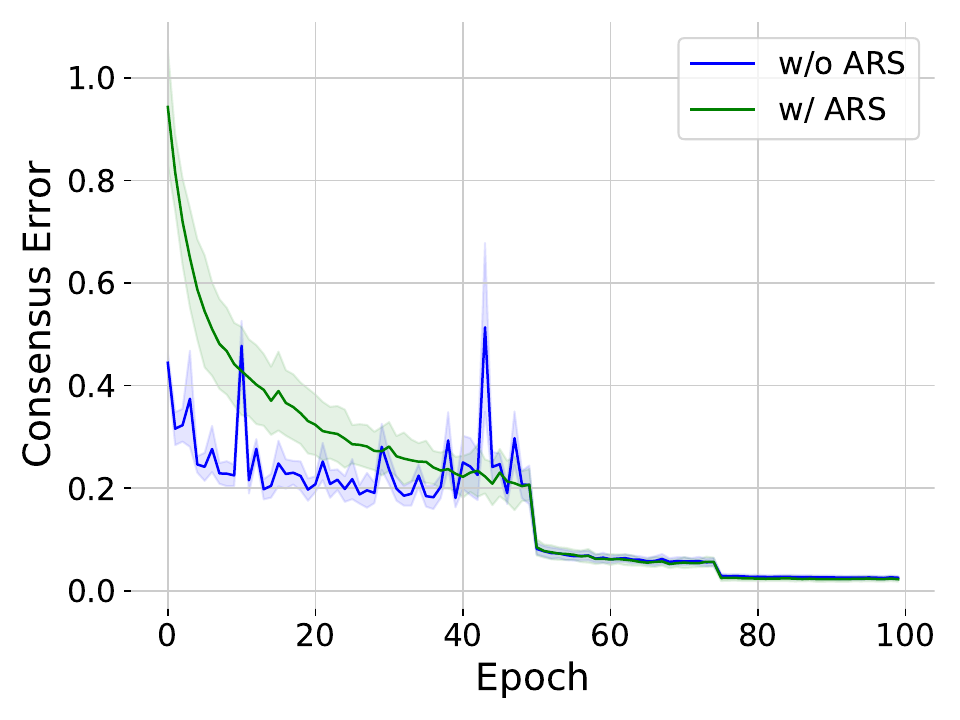}\label{fig:ce_fm}}
  \caption{Average consensus error during training for various datasets trained over ring topology of 16 agents. The graph is plotted for one seed where the solid line represents the average consensus error across the agents and the shaded region represents the variation of the consensus error across agents.}
  \label{fig:ce}
\end{figure}

\begin{figure}
  \centering
  \subfloat[CIFAR-10 trained on a 5-layered CNN]{\includegraphics[width=0.4\textwidth]{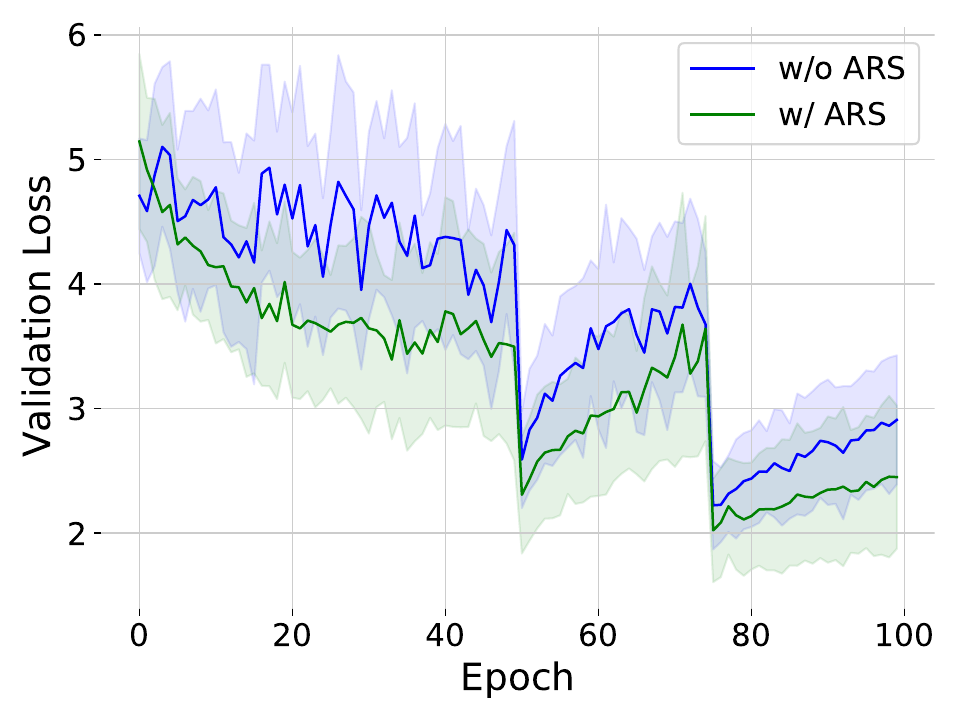}\label{fig:loss_cf10}}
  \hspace{4mm}
  \subfloat[Fashion-MNIST trained on LeNet-5]{\includegraphics[width=0.4\textwidth]{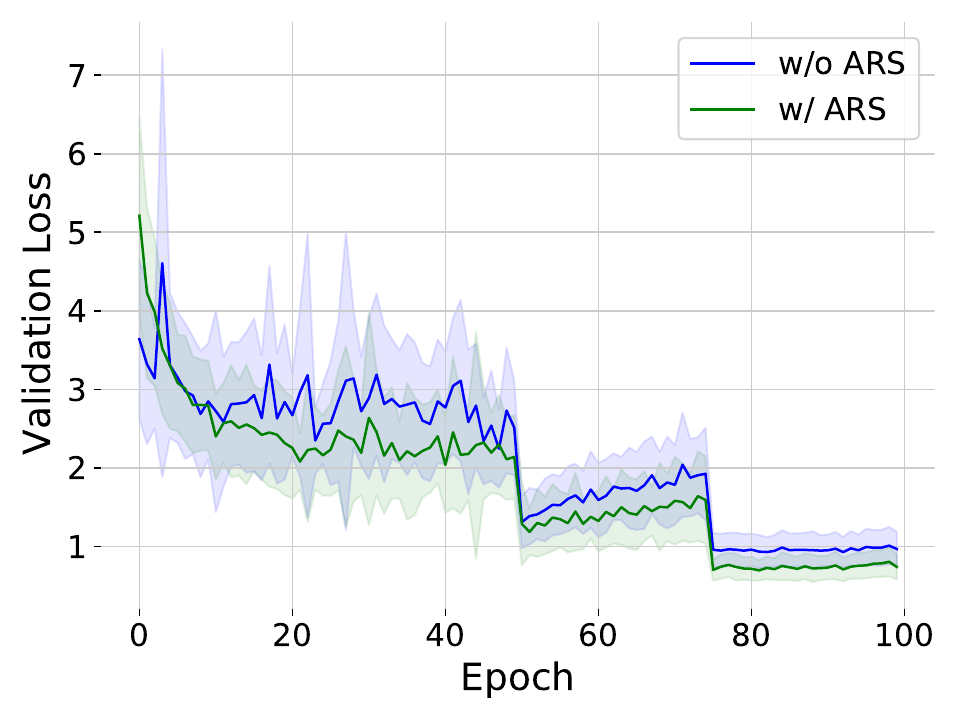}\label{fig:loss_fm}}
  \caption{Average validation during training for various datasets trained over ring topology of 16 agents. The graph is plotted for one seed where the solid line represents the average loss across the agents and the shaded region represents the variation of the loss across agents.}
  \label{fig:loss}
\end{figure}
Table.~\ref{table:datasets} presents the results of various datasets trained in a decentralized manner using the D-PSGD algorithm with and without ARS. We train the Fashion MNIST dataset on LeNet-5 architecture, CIFAR-100 on ResNet-20, and ImageNette on MobileNet-V2. We observe an average improvement of $2\%$ in test accuracy across various datasets.

\begin{table} [h]
\begin{minipage}{0.46\linewidth}
		\caption{Test accuracy of CIFAR-10 dataset trained on ResNet-20 distributed over a ring topology with a skew of $\alpha=0.01$.}
		\label{table:agents}
		\centering
		\begin{tabular}{lcc}
			\toprule
			Agents &   w/o ARS & w/ ARS\\
			\midrule
            16 & $67.67 \pm 4.48$ & $\mathbf{71.19} \pm 3.66$ \\
            24 & $ 65.98 \pm 3.64$ & $\mathbf{69.91} \pm 3.45$ \\
            32 & $61.65 \pm 3.51$ & $\mathbf{68.58} \pm 1.77$ \\
             48 & $56.56 \pm 11.52$ & $\mathbf{61.79} \pm 6.28$ \\
			\bottomrule
		\end{tabular}
	\end{minipage} \hfill
	\begin{minipage}{0.52\linewidth}
		\caption{Test accuracy of various datasets trained over a ring topology of 16 agents with a skew of $\alpha=0.01$.}
		\label{table:datasets}
		\centering
		\begin{tabular}{lcc}
			\toprule
			Dataset  & \multirow{2}{*}{w/o ARS} &\multirow{2}{*}{ w/ ARS}\\ 
   (Model) &  & \\ 
			\midrule
            Fashion MNIST & \multirow{2}{*}{$84.58 \pm 0.32$} & \multirow{2}{*}{$ \mathbf{85.55} \pm 0.81$} \\
            (LeNet-5) & &  \\
            \hline
            CIFAR-100 & \multirow{2}{*}{$41.57 \pm 0.72$} & \multirow{2}{*}{$ \mathbf{44.05} \pm 1.70$} \\ 
            (ResNet-20) &  &  \\ 
             \hline
            ImageNette & \multirow{2}{*}{$43.68 \pm 8.16$} & \multirow{2}{*}{$ \mathbf{46.59} \pm 4.73$} \\ 
            (MobileNet-V2) &  &  \\ 
			\bottomrule
		\end{tabular}
	\end{minipage}
\end{table}

Additionally, we compute the average consensus error i.e., $\frac{1}{n} \sum_i^n ||x_i^t-\Bar{x}||_F^2$ over time for various datasets trained over ring topology of 16 agents. $\Bar{x}$ is the average/consensus model parameters. We observe that at the end of the training, the consensus error with or without ARS is similar (refer Fig.~\ref{fig:ce}). 
The main difference is during the initial epochs of the training, where the consensus error for D-PSGD with ARS gradually decreases over time in contrast to D-PSGD without ARS which oscillates at a lower value.
Note that the neighborhood's model information is not completely reliable at the initial phase of training as the models are just starting to learn and the variation in parameters across agents is high because of data heterogeneity.
Since ARS starts with a lower averaging rate and exponentially increases it while training, it avoids aggressive averaging with the neighborhood's models at the initial epochs.  This in turn helps in improving the performance (reducing the validation loss) of decentralized learning on heterogeneous data as shown in Fig.~\ref{fig:loss}. 

\subsection{Limitations and Future Work}
\label{apx:future}

The inherent limitation of the proposed method is the hyper-parameter tuning of the initial averaging rate, growth rate, type of decay, etc. Similar to the learning rate scheduler, one needs to curate the ARS for the given dataset, model architecture, graph structure, and data distribution. 

There are two potential future directions for this work:

1. A study on designing and evaluating the impacts of ARS on algorithms such as QGM \citep{qgm}, GUT \citep{gut}, and (NGM\textsubscript{mv}) \citep{ngm} is a potential extension for this work. 
Quasi Global Momentum (QGM), Global Update Tracking (GUT), and Neighborhood Gradient Mean (NGM\textsubscript{mv}) are well-known and current state-of-the-art methods for decentralized learning with heterogeneous data that do not require any communication overhead. These methods use additional memory and compute resources to improve the test accuracy and perform much better than D-PSGD and its ARS variant. ARS is a simple hyper-parameter scheduling method and can not independently compete with methods that modify the gradient or momentum information to improve performance.
However, all these algorithms (QGM, GUT, and NGM\textsubscript{mv}) also utilize the averaging rate hyper-parameter. Therefore, ARS can be employed in synergy with these methods to further improve the performance of decentralized learning on heterogeneous data.

2. Theoretical analysis of the convergence rate of the D-PSGD algorithm with an Averaging Rate Scheduler (ARS) can help in better understanding the impact of the ARS. Analyzing the convergence rate with ARS  can shed light on whether ARS leads to faster convergence to the same solution or allows D-PSGD to converge to a better stationary point of the global objective. Note that the existing theoretical analysis of D-PSGD \citep{d-psgd} assumes a constant averaging rate, leaving room for potential future work.

\end{document}